\documentclass{article}
\usepackage{amsfonts}
\usepackage[cmex10]{amsmath}

\usepackage{times}
\usepackage{cite}
\usepackage{amsmath}
\onecolumn
\usepackage{graphicx}
\usepackage{latexsym}
\usepackage{subfigure}
\usepackage{pstricks}
\usepackage{setspace}
\date{}
\begin{document}
\title{WEBCA: Weakly-Electric-Fish Bioinspired Cognitive Architecture}
\author{Amit Kumar Mishra \\
University of Cape Town\\
akmishra@ieee.org}

%\footnotetext[1]{This paper is presented in the Biologically Inspired Cognitive Architecture Conference 2017 and published by their proceedings.}
% make the title area

\maketitle
\begin{abstract}
Neuroethology has been an active field of study for more than a century now. Out of some of the most interesting species that has been studied so far, weakly electric fish is a fascinating one. It performs communication, echo-location and inter-species detection efficiently with an interesting configuration of sensors, neurons and a  simple brain. In this paper we propose a cognitive architecture inspired by the way these fishes handle and process information. We believe that it is easier to understand and mimic the neural architectures of a simpler species than that of human. Hence, the proposed architecture is expected to both help research in cognitive robotics and also help understand more complicated brains like that of human beings.  
\end{abstract}

\section{Introduction}
%jamming avoidance response (JAR)
%electrosensory lateral line lobe (ELL): the only nucleus that receives direct input from peripheral electroreceptor afferents.
%electric organ dis- charges (EODs)
%Amplitude encoding afferent fibers in wave gymnoti- forms are referred to as P-type (probability-type) afferents. 
 
Bio-inspired cognitive architectures (BICA) have been active fields of research for more than a decade. BICAs help us in both understanding (and modelling) brain actions and in trying to implement computations that may, potentially, mimic intelligent behaviour. 

Cognitive architectures like SOAR and ACT-R \cite{rose_91_soar, taat_06_actr} have been pretty successful. However incorporating them into real system brings the symbolic-nonsymbolic fusion challenge \cite{sam_10_bica, lai_08, lang_09, kel_09}. 
That is why there are not many works in implementing BICA for a real system like a robot. 
Only recently there have been some  progress in this front \cite{cogrob_nature}. 

In a pursuit to have a BICA which, thought limited in its performance,  might be easier to implement we looked into other members of biosphere. The aim was to study the cognitive architecture of an animal which is simple and well studied; and yet shows remarkable performance which are not easily achievable by current generation algorithms and robots. 
We decided to focus on weakly electric fish (WEF). Its neural processing is relatively well studied and documented \cite{bell_89, brian_00, nelson_02, swat_05, walter_87, krahe_14, car_86} including some attempts at generating detailed inverse problem based model for the fish \cite{ammari_16} and robotics platform inspired by the fish \cite{iver_04}. Also the fish shows some remarkable performance especially in terms of achieving hyperacuity \cite{walter_87}. 
This made us hypothize that it might be more practical to focus on neuroethology to incorporate limited cognitive abilities into practical systems \cite{brian_00}. 
In the current paper, we propose a BICA inspired by the way weakly electric fish  process their information. We call it WEBCA (Weakly-Electric-Fish Bioinspired Cognitive Architecture).
We believe that the proposed architecture, because of its simplicity, will be easy to implement and can help us to incorporate limited cognitive abilities into a system like a robot.

Rest of the paper is detailed as follows. Section 2 gives some of the salient aspects of  neuroethology of weakly electric fishes (WEF). In Section 3 we describe WEBCA. 
We conclude our work and discuss future directions in Section 4.

\section{Neuroethology of Weakly Electric Fish}
Weakly electric fishes are simple creatures with some sophisticated abilities. These fishes, mostly found in deep rivers (with very little light), use electricity for three major functions; viz. passive electro-location (by sensing the electric signal emitted by almost all other animals around); active electro-location (by sensing the reflection of electric signal generated by their electric organ discharge (EOD)); and electro-communication (with different members of their species). 
There are different sensors in the skin of the fish for these three actions. These sensory cells work in different ways to collect these information. 
For example the sensory cells responsible for electro-communication (Knollenorgans Electroreceptors) are phase receptors and respond to transients \cite{bell_89}. 

The receptors end in  the electrosensory lateral line lobe (ELL). Something interesting starts happening now. Each electroreceptor trifurcates to three ELL maps: the centromedial (CMS), centrolateral (CLS) and lateral (LS) segments. 
These three are of different sizes (with CMS being the biggest and LS being the smallest map (In terms of neurone assigned)). 
They also process the signal for different features so that different behaviours are mapped to different types of burst characteristics. Another interesting phenomena is the fact that the frequency response of the CLS cells changes as per the context (e.g. communication or foraging). 
 This multiple parallel mapping of information is very similar to the way most convolutional neural network (CNN) work. 
 The big difference, however, lies in the way these layers are tuneable depending on the context for the fish. 
 
 The ELL signal are next mapped to  torus semicircularis (TS) where the information is further organised to many more layers (around 12) and more than 50 neurone types. The exact working of these layers are not very well understood. However, recent studies do show that these are linked with very fine feature extraction.  
 
 Information from TS goes to tectum (which is equivalent to the visual cortex of mammals). Both tectum and TS feed information to pallium (which is similar to cortex of mammals). Pallium can, in turn, affect the way TS and tectum map sensory information. The detailed block diagram is given in Figure 1 taken from \cite{krahe_14}.

\begin{figure}[htbp]
\includegraphics [scale=0.9]{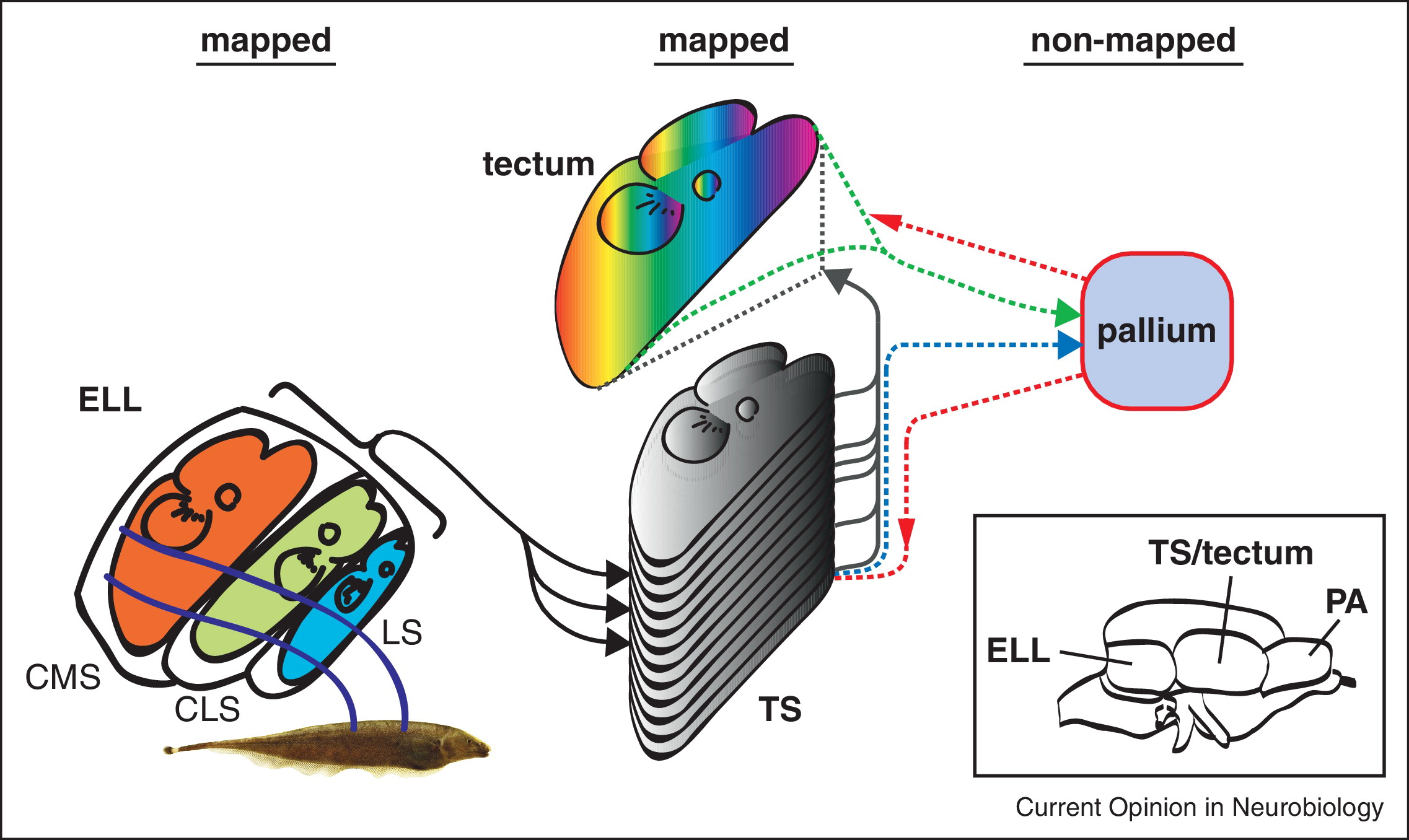}
\label{orig}
\caption{Mapping of sensory information in weakly electric fish \cite{krahe_14}.}
\end{figure}

A WEF, like many animals, can achieve a range of marvellous tasks. We shall focus on the following  tasks which we would try to keep in mind while designing our BICA. 
\begin{enumerate}
\item {\bf Coding based sensor tuning:} Depending on its situation and motif the fish has the ability to tune what information is collected and represented at the sensors. This is changed by deciding what coding scheme shall be used by the sensors in the skin to send across to  ELL \cite{swat_05}. This is an elegant way to fuse symbolic and non-symbolic layers in modern AI. 
It is not the best solution (as its not very generic). However, it can be the starting point. 
\item {\bf Hyper-acuity:} Many animals can resolve events at a scale which is finer than what their sensors can achieve. 
The WEF's neurological study seems to suggest that this is achieved in the fish by the use of multiple spatial representation of the information coming from the sensors. 
And the burst characteristics in these maps represent various events \cite{nelson_02, swat_05}. 
Hence it seems the ability of apparently higher resolution of detecting ``certain events'' comes from blowing up the information into a spatial map and inferring from these maps. 
\item {\bf Ambiguity of information from a single neuron: } Multiple spatial neural maps are the way to handle and extract information. A single neuron will be part of multiple such mappings. Hence, the exact information represented by a single neuron is ambiguous. This helps in making the whole system more robust and less dependent on local blocks. 
\end{enumerate}

\section{Weakly-Electric-Fish Bioinspired Cognitive Architecture \label{arch}}
Following our discussion in the section before, we now present the BICA that we have derived inspired by WEF. 
The architecture is represented in Figure 2.

\begin{figure}[htbp]
\includegraphics [scale=0.5]{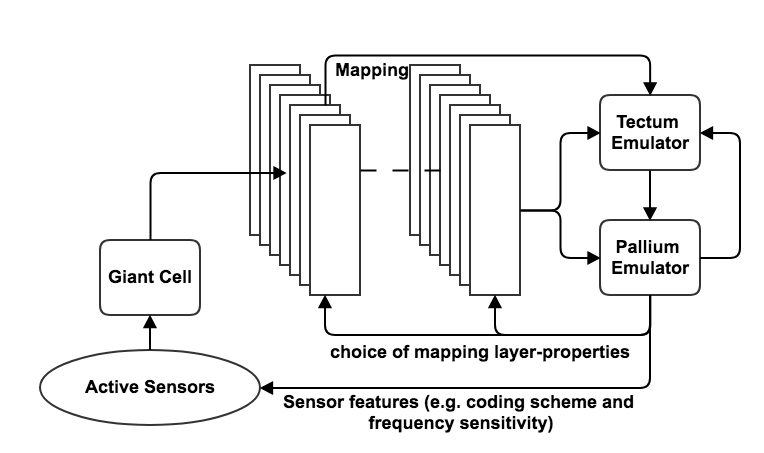}
\label{weba}
\caption{Block level presentation of WEBA.}
\end{figure}

We shall now, briefly, describe the major blocks and the way these can be implemented. 
It can be noted that there might be multiple (and more elegant) ways to implement these blocks. 

The sensors are called active because they can  be, actively, emitting their own signal and recording the echo. 
Another meaning of active would be that the sensor will  have the ability to change how it behaves. For example even if its a light sensor (which is passive by conventional definition), if it has the ability to change what information it sends up (and the ability to change it) it can be called active. 
\\
{\bf Definition} An active sensor is one which either transmits its own energy or has the ability to change information representation in-situ. 
\\
The way we want to implement this is by using a wavelet based sensing. Immediately after the sensing the signal can be passed through a wavelet based representation. The mother wavelet shall be decided by the Pallium Emulator. 

The mapping is well modelled by a convolutional neural network. 
The trick, however, shall lie in how to implement the tuning of mapping layer properties on the fly. For this we intend to use dynamic graphs (which are implementable with ease using the new features of PyTorch. 

The Tectum Emulator is like the fully connected layer of conventional deep neural networks. 
It extracts symbolic information from the multi-level maps. 
The main difference in our architecture will be the fact that this block gathers spatial information from multiple layers. 

The Pallium Emulator  is a symbolic computational block which gathers information from the Tecturm Emulator and also from the layers of neural networks. 
A WEF deals with three major types of tasks (active-echolocation, passive-echolocation and communication) and hence can change the mapping and sensor parameters accordingly. In our case we can set a broad range of possible situations and accordingly let the Pallium Emulator know what action to be taken. Accordingly it can send the necessary command to the mapping layers and to the active sensors.

%\section{Suggested Test-cases}

%\begin{enumerate}
%\item {\bf Westcott test based test-case:} 
%\end{enumerate}

\section{Conclusion}
 In this paper we presented an cognitive architecture inspired by the neuroethology of weakly electric fish. 
 The fact that the neurological action of the fish has been well studied was one of the factors which inspired us. In addition to this we believed that it might be better (and easier) to model lower vertebrates than directly aiming to model human brain which is much more complicated and much less understood. 
 In our architecture we tried to focus on two major aspects of the fish which we considered to be interesting to be emulated. 
 The first of these aspects, which we believe our architecture shall achieve, is global feedback. 
 Global feedback has been one of the proposed schemes to connect symbolic layers with non-symbolic layers. The fish manages this by controlling the modulation and sensitivity of its sensors. 
 The second aspect, we aimed at, was hyper-acuity. The fish processes the information by representing representing the stimulus as a neural map which is believed to be a major reason for how it achieves hyper-acuity. 
 
 In our future work we shall implement this architecture on a robotic platform and validate its performance. 
 %social media comment analysis using our system and validate its working. 
 
\bibliographystyle{IEEEtran}
\bibliography{ref}

%\begin{figure}[htbp]
%\includegraphics [scale=.4]{FIG4.eps}
%\caption{Membership functions before Training FIS }
%\end{figure}
%%\end{document}\\
%
%\begin{figure}[htbp]
%\includegraphics [scale=.4]{FIG5.eps}
%\caption{Membership functions after training FIS}
%\end{figure}

\end{document}